\documentclass[10pt,twocolumn,letterpaper]{article}

\usepackage{icb}
\usepackage{times}
\usepackage{epsfig}
\usepackage{graphicx}
\usepackage{amsmath}
\usepackage{amssymb}
\usepackage{subcaption}
\usepackage{color}
\usepackage{amsfonts}
\usepackage{float}
\usepackage{multirow}
\usepackage[linesnumbered,ruled,vlined]{algorithm2e}
\usepackage{textcomp}
\usepackage{slashbox}
\usepackage{makecell}

\icbfinalcopy 

\ificbfinal\fi
\begin{document}

\title{Fully Associative Patch-based 1-to-N Matcher for Face Recognition}

\author{Lingfeng Zhang and Ioannis A. Kakadiaris\\
	Computational Biomedicine Lab \\
	University of Houston\\
	{\tt\small \{lzhang34, ioannisk\}@uh.edu}
}
%

\maketitle
\thispagestyle{empty}

\begin{abstract}
This paper focuses on improving face recognition performance by a patch-based 1-to-N signature matcher that learns correlations between different facial patches. A Fully Associative Patch-based Signature Matcher (FAPSM) is proposed so that the local matching identity of each patch contributes to the global matching identities of all the patches. The proposed matcher consists of three steps. First, based on the signature, the local matching identity and the corresponding matching score of each patch are computed. Then, a fully associative weight matrix is learned to obtain the global matching identities and scores of all the patches. At last, the $\ell_1$-regularized weighting is applied to combine the global matching identity of each patch and obtain a final matching identity. The proposed matcher has been integrated with the UR2D system for evaluation. The experimental results indicate that the proposed matcher achieves better performance than the current UR2D system. The Rank-1 accuracy is improved significantly by 3\% and 0.55\% on the UHDB31 dataset and the IJB-A dataset, respectively.
\end{abstract}


\section{Introduction}

Face recognition is one of the major visual recognition tasks in the fields of biometrics, computer vision, image processing, and machine learning. In recent years, most of the significant advances in visual recognition have been achieved by deep learning models, especially deep Convolutional Neural Networks (CNNs) \cite{ILSVRC2015, girshick2016region, szegedy2013deep}. CNN was first proposed in the late 1990s by LeCun \textit{et al.} \cite{lecun1989backpropagation, lecunhandwritten}, but was quickly overwhelmed by the combination of other shallow descriptors (such as SIFT, HOG, bag of words) with Support Vector Machines (SVMs). With the increase of image recognition data size and computation power, CNN has become more and more popular and dominant in the last five years. Krizhevsky \textit{et al.} \cite{Alex2012} proposed the classic eight-layer CNN model (AlexNet) with five convolutional and three fully connected layers. The model is trained via back-propagation through layers and performs extremely well in domains with a large amount of training data. Since then, many new CNN models have been constructed with larger sizes and different architectures to improve performance. Simonyan \textit{et al.} \cite{simonyan2014very} explored the influence of CNN depth by an architecture with small convolutional filters ($3 \times 3$). They achieved a significant improvement by pushing the depth to 16-19 layers in a VGG model. Szegedy \textit{et al.} \cite{szegedy2015going} introduced GoogLeNet as a 22-layer Inception network, which achieved impressive results in both image classification and object detection tasks. He \textit{et al.} \cite{He_2016_CVPR} proposed Residual Network (ResNet) with a depth of up to 152 layers, which set new records for many image recognition tasks. Furthermore, He \textit{et al.} \cite{he2016identity2} released a residual network of 1,000 layers with a new residual unit that makes training easier and improves generalization.
 
In the recent years, many CNNs have been introduced in face recognition and have achieved a series of breakthroughs. Similar to image recognition, effective CNNs require a larger amount of training images and larger network sizes \cite{zhang2017ijcb}. Yaniv \textit{et al.} \cite{taigman2014deepface} trained the DeepFace system with a standard eight-layer CNN using 4.4M labeled face images. Sun \textit{et al.}  \cite{sun2014deep, sun2014deep2, sun2015deepid3} developed the Deep-ID systems with more elaborate network architectures and fewer training face images, which achieved better performance than the DeepFace system. FaceNet \cite{schroff2015facenet} was introduced with 22 layers based on the Inception network \cite{szegedy2015going, zeiler2014visualizing}. It was trained on 200M face images and achieved further improvement. Parkhi \textit{et al.} \cite{parkhi2015deep} introduced the VGG-Face network with up to 19 layers adapted from \cite{simonyan2014very}, which was trained on 2.6M images. This network also achieved comparable results and has been extended to other applications. To overcome the massive request of labeled training data, Masi \textit{et al.} \cite{masi16dowe} proposed to use domain-specific data augmentation, which generates synthesis images for the CASIA WebFace collection \cite{yi2014learning} based on different facial appearance variations. Their results trained with ResNet match the state-of-the-art results reported by the networks trained on millions of images. 

In order to overcome pose variations, Xu \textit{et al.} \cite{xiang2017ijcb} presented the evaluation of a pose-invariant 3D-aided 2D face recognition system (UR2D) that is robust to pose variations as large as 90\textdegree{}. Different CNNs are integrated in face detection, landmark detection, 3D reconstruction, and signature generation. The texture-lifted image is divided into multiple patches for signature generation. Then, the similarity scores of all the patches are combined with occlusion information to obtain a final similarity score for each pair of gallery and probe matching. The rank-1 matching identity of a probe is obtained with the maximum similarity score over the whole gallery in 1-to-N matching. However, there are two major limitations in the current matcher of UR2D system: (a) the facial patches are considered separately, the correlations between these patches are ignored; (b) the similarity scores of all the patches are added directly, the differences between these patches are neglected. 

\begin{figure*} 
	\centering
	\begin{center}
		\includegraphics[width=1\linewidth]{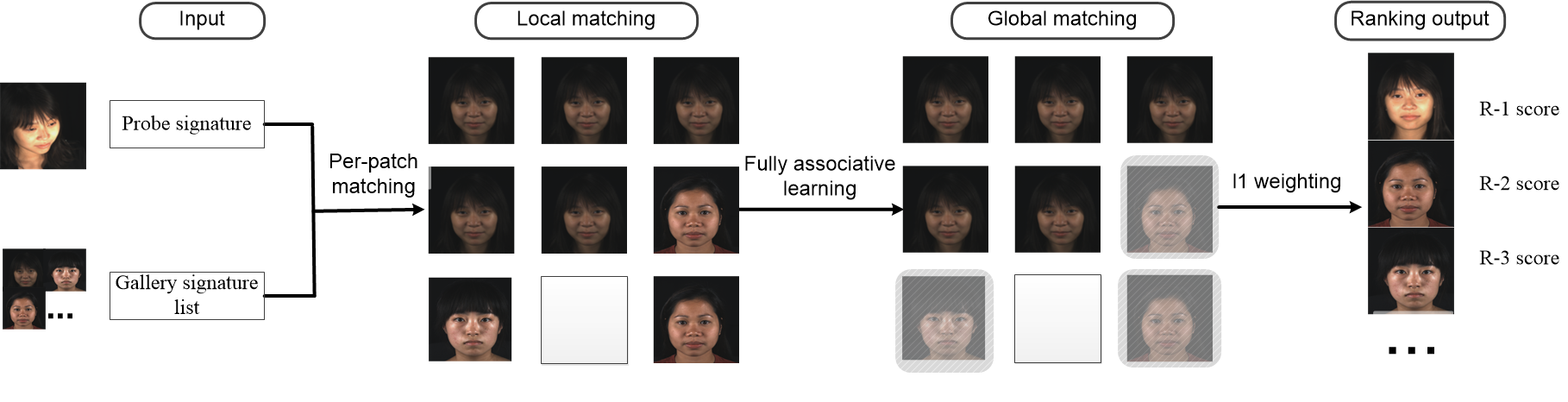}
	\end{center}
	\caption{Depicted the procedure of the proposed patch-based 1-to-N matcher.}
	\label{pipeline0}
\end{figure*}

This paper overcomes the limitations (a) and (b) by introducing patch correlation learning in signature matching. The two terms ``local matching" and ``global matching" are used to refer the matching result before and after fully associative learning, respectively. Given a probe signature and gallery signature list, the local matching identity and score of each patch are first computed based its patch signature. Then, a fully associative weight matrix is learned to update the local matching identity and score of each patch and obtain the global matching identity and score. The term ``fully associative" is derived from the fact that each patch's local matching identity has its contributions to the global matching identities of all the patches. This way if one patch is occluded or mismatched, its matching score will be decreased by using the fully associative weight matrix. Meanwhile, the correct matching patches will be boosted. Furthermore, the $\ell_1$-regularized weighting is applied to combine the global matching identity of each patch rather than directly summing up the similarity score of each patch. The procedure of the proposed patch-based matcher is shown in Figure~\ref{pipeline0}.

The contribution of this paper is improving face recognition signature matching by introducing two techniques: (i) actively associating the patch correlations by proposing a fully associative model to learn the correlations of matching results between different facial patches. The fully associative weight matrix is learned based on the kernel trick. (ii) improving the final matching accuracy by introducing the $\ell_1$-regularized weighting-based method to combine the global matching results of all the patches.

The rest of this paper is organized as follows: Section \ref{sec2} presents the related work. Section \ref{sec3} describes the signature. Section \ref{sec4} introduces the proposed signature matcher. The experimental design, results, and analysis are presented in Section \ref{sec5}. Section \ref{sec6} concludes the paper. 

\section{Related work}
\label{sec2}
 In the history of face recognition, both global and local methods have been developed. Global methods learn discriminative information from the whole face image, such as subspace methods \cite{turk1991eigenfaces, belhumeur1997eigenfaces}, Sparse Representation based Classification (SRC) \cite{wright2009robust,yang2011robust} and Collaborative Representation based Classification (CRC) \cite{zhu2012multi,zhang2011sparse}. Although global methods have achieved great success in controlled environments, they are sensitive to the variations of facial expression, illumination and occlusion in uncontrolled real-world scenarios. Proven to be more robust, local methods extract features from local regions. The classic local features include Local Binary Patterns (LBP) \cite{ahonen2006face, liao2007learning}, Gabor features \cite{zhang2005local, su2009hierarchical}, Scale-Invariant Feature Transform (SIFT) \cite{luo2007person, bicego2006use}, gray values, and CNN features. In local methods, more and more efforts focus on patch (block) based methods, which usually involve steps of local patch partition, local feature extraction, and local prediction combination. With intelligent combination, these methods weaken the influence of variant-prone or occluded patches and combine the prediction of invariant or unoccluded patches. Martinez \cite{martinez2002recognizing} proposed to divide face images into several local patches and model each patch with a Gaussian distribution. The final prediction is reached by summing the Mahalanobis distance of each patch. Wright \textit{et al.} \cite{wright2009robust} extended SRC into a  patch version that achieved better performance by a voting ensemble. Taking into account the global holistic features, Su \textit{et al.} \cite{su2009hierarchical} developed a hierarchical method that combines both global and local classifiers. Fisher linear discriminant classifiers are applied to global Fourier transform features and local Gabor wavelet features. A two-layer ensemble is proposed to obtain the final prediction. To overcome the impact of patch scale, Yuk \textit{et al.} \cite{yuk2011multi} proposed a Multi-Level Supporting scheme (MLS) with multi-scale patches. First, Fisherface based classifiers are built on multi-scale patches. Then, a criteria-based class candidate selection technique is designed to fuse local prediction. Zhu \textit{et al.} \cite{zhu2012multi} developed Patch-based CRC (PCRC) and Multi-scale PCRC (MPCRC). The constrained $\ell_1$-regularization is applied to combine each patch's local prediction. Zhang \textit{et al.} \cite{zhang2015icb} developed a patch-based hierarchical multi-label method for face recognition. A face image is divided into multi-level patches iteratively and labeled with hierarchical labels. The hierarchical relationships defined between local patches are used to obtain the global prediction of each patch. The drawback of most previous patch-based methods is that they rely on shallow features, rather than deep features.  
 
 The proposed method is also related to the Hierarchical Multi-label Classification (HMC) problem, where each sample has more than one label and all these labels are organized hierarchically in a tree or Direct Acyclic Graph (DAG) \cite{silla2011survey}. Label correlations in tree and DAG structures are used to improve classification performance \cite{valentini2011true,zhang2014fully}. However, these methods were developed for multi-label classification problems, rather than single label matching.

\section{Signature}
\label{sec3}
Given an input face image, the pipeline of UR2D follows: face detection, landmark detection, pose estimation, 3D reconstruction, texture lifting, signature generation, and signature matching. Here focuses on the signature generation part, please refer to Xu \textit{et al.} \cite{xiang2017ijcb} for more details. The signature of each face image is extracted from its texture-lifted image. Facial texture lifting is a technique that lifts the pixel values from the original 2D images to a UV map \cite{Kakadiaris2017137}. Given an original image, a 3D-2D projection matrix \cite{dou2015pose}, a 3D AFM model \cite{Kakadiaris2017}, it first generates the geometry image, each pixel of which captures the information of an existing or interpolated vertex on the 3D AFM surface. With the geometry image, a set of 2D coordinates referring to the pixels on an original 2D facial image is computed. Thus, the facial appearance is lifted and represented into a new texture image. The 3D model and a Z-Buffer technique are applied to estimate the occlusion status for each pixel. This process also generates an occlusion mask. 

\begin{figure} 
	\centering
	\begin{center}
		\includegraphics[width=1\linewidth]{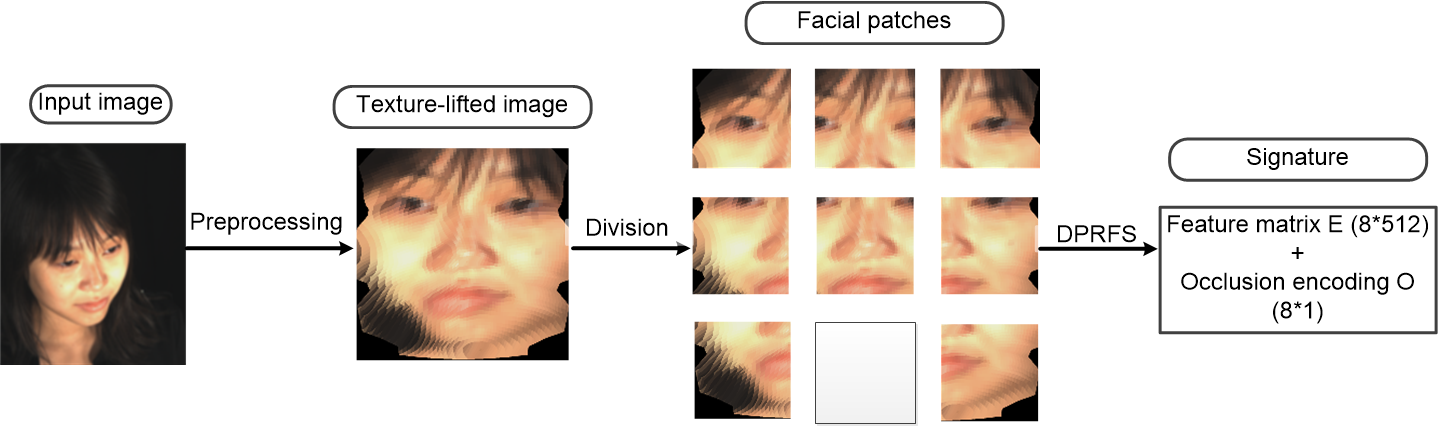}
	\end{center}
	\caption{Depicted the procedure of DPRFS-based signature generation.}
	\label{pipeline1}
\end{figure}

In UR2D, two types of signatures can be extracted \cite{dou2015pose}： Pose Robust Face Signature (PRFS) and Deep Pose Robust Face Signature (DPRFS). Both PRFS and DPRFS are patch-based signature. In PRFS, the facial texture image and the self-occlusion mask image are first divided into 64 non-overlapping local patches. Then, on each local patch, the discriminative DFD features \cite{lei2014learning} are extracted. Also a self-occlusion encoding is computed. Based on CNN features, DPRFS achieves better performance. In DPRFS, the facial texture image and the self-occlusion mask image are first divided into eight partially-overlapping local patches. The mouth patch is ignored due to expression variations. Then, a ResNet model is trained for each patch based on softmax loss and center loss. The signature consists of two parts: feature matrix and occlusion encoding. Figure~\ref{pipeline1} depicts the procedure of signature generation based on DPRFS. Let $E = \{ e_{ij} \}^{b\times m} = \{E_1, E_2, ..., E_m\}$ represent a feature matrix, where each value $e_{ij}$ represent the $i^{th}$ feature of the $j^{th}$ patch while $b$ and $m$ represent the number of features and the number of patches, respectively. The feature vector of the $i^{th}$ patch is presented by $E_i$. The occlusion encoding is represented by $O = \{o_1, o_2, ...,o_m\}$, where $o_j$ is a binary value indicating whether the $j^{th}$ patch is non-occluded. Based on the occlusion encoding of each patch, all the feature vectors are combined selectively during matching. Let $\mathbb{S}={\{E}, {O}\}$ represent the signature based on DPRFS. The $\mathbb{S}$ signature size is $8 \times 512 + 8$. The same preprocessing and DPRFS signature generating process are followed as the UR2D system, summarized in Algorithm~\ref{a1}. Improving the signature matching process will be introduced in the next section. 

\begin{algorithm}
	\caption{Signature: $\mathbb{S}$}
	\label{a1}
	\KwIn{ 2D image $I$ and 3D AFM model $M$}
	\KwOut{$\mathbb{S}=\{E, O\}$}
	Face detection and landmark detection \\
	Pose estimation and 3D reconstruction \\	
	Generate geometry image  \\
	Compute texture lifted image and occlusion mask\\
	Compute feature matrix $E$ and occlusion encoding $O$ \\
	\Return{\{$\mathbb{S}=\{E, O\}$\}};
\end{algorithm}

\section{Fully associative patch-based matcher}
\label{sec4}
In the UR2D system, The cosine score is used to measure the similarity between different feature vectors. Let $I^g$ and $I^p$ represent a pair of gallery image and probe image. Their feature matrix and occlusion encoding are represented by $E^g$, $E^p$ and $O^g$, $O^p$, respectively. The feature is patch based, and only non-occluded patches contribute to $s$. The signature matching score $s$ is computed as:
\begin{equation}\label{sm2}
s=\frac{1}{k}\sum\limits_{j=1}^{m}(o^g_j \& o^p_j)\times cosine(E^g_j, E^p_j),  \\
\end{equation} 
where $k$ represents the number of the non-occluded patch pairs. Then, the identity with the maximum score of the whole gallery is return as the 1-to-N matching result. The limitation is that the correlations between different patches are neglected. Also, all the patches are treated equally.

In this section, the idea of fully associative learning is introduced into patch-based face recognition by making use of the relationships between different patches. The proposed matcher consists of three major steps. First, local matching: the local matching identity and corresponding score of each patch are computed based on its feature vector in signature. Second, fully associative learning: a weight matrix is learned to improve the local matching identity and score of each patch and obtain its global matching identity and corresponding score. Third, global matching combination: the $\ell_1$-regularized weighting is applied to combine the global matching identity of each patch. 

\subsection{Fully associative learning}
In the proposed matcher, the local matching identity and corresponding matching score of each patch are obtained firstly and used to learn the correlations between different patches. Assume the identity set of gallery and probe set used in training is represented by $\mathcal{L}=\{1,2,...,l\}$. Let $\mathcal{S}=\{s_1,s_2,...,s_n\}$ represent probe set, which comprises $n$ samples. Its identity vector is denoted by  $\mathcal{C}=\{c_1,c_2,...,c_n\}$, where $c_i \in \mathcal{L}$. Its local matching identity matrix is defined as a $P=\{p_{ij}\}$, with size $n\times m$, where each element $p_{ij}$ represents the matching identity of the $i^{th}$ sample's $j^{th}$ patch. So $p_{ij} \in \mathcal{L}$. Also, the local matching score matrix is defined as a $Z=\{z_{ij}\}$, with size $n\times m$, where each element $z_{ij}$ represents the corresponding matching score of each local matching identity. Each score value is computed from the cosine score and $z_{ij} \in [0,1]$. If one patch is occluded, the score is set to 0. A corrected local matching score matrix is defined as $D=\{d_{ij}\}$ with the same size as $Z$. The value of $d_{ij}$ is binary and decided by whether the local matching of $p_{ij}$ is correct or not:
\begin{equation}\label{out1}
d_{ij} = \left\{ \begin{array}{rcl}
1  & p_{ij} = c_i \\
0  & p_{ij} \neq c_i \\
\end{array}\right..
\end{equation}

Let $Y=\{y_{ij}\}$ represent the global matching score matrix based on fully associative learning. To take all the patch-to-patch relationships into account, $W=\{w_{ij}\}$ is defined as a weight matrix, where $w_{ij}$ represents the weight of the $i^{th}$ patch's local matching score to the $j^{th}$ patch's global matching score. Thus, each patch's global matching score is a weighted sum of the local matching scores of all the patches. The global matching matrix $Y$ is computed as: $Y=ZW$.

The simplest way to estimate the weight matrix $W$ is by minimizing the squared loss between the global matching score matrix $Y$ with the corrected local matching matrix $D$. To reduce the variance of $w_{ij}$, the Frobenius norm of $W$ is included which leads this objective function:

\begin{equation}\label{op1}
\min_{W}\|D-ZW\|_F^2+\lambda_1\|W\|_F^2,
\end{equation}
where the first term measures the empirical loss of the probe set, the second term controls the generalization error, and $\lambda_1$ is a regularization parameter. The above function is known as ridge regression. Taking derivatives w.r.t. $W$ and setting to zero, the solution is:

\begin{equation}\label{w1}
W=\left(Z^TZ+\lambda_1I_m\right)^{-1}Z^TD,
\end{equation}
where $I_m$ represents the $m\times m$ identity matrix. Thus, an analytical solution is obtained for the weight matrix. 

To capture the complex correlation between global and local matching score, the above formula is generalized using the kernel trick. Let $\Phi$ represent the map applied to each sample's local matching vector $\mathbf{z}_i$. A kernel function is induced by $K(\mathbf{z}_i,\mathbf{z}_j)=\Phi(\mathbf{z}_i)^T\Phi(\mathbf{z}_j)$. By replacing the term $Z$ in (\ref{op1}):

\begin{equation}\label{op2}
\min_{W_{k}}\|D-\Phi W_{k}\|_F^2+\lambda_1\|W_{k}\|_F^2.
\end{equation}
After several matrix manipulations \cite{an2007face}, the solution of  $W_{k}$ becomes:

\begin{equation}\label{wk}
\begin{split} 
W_{k} & =\left(\Phi^T\Phi+\lambda_1I_l\right)^{-1}\Phi^TD \\
&   =\Phi^T \left(\Phi\Phi^T+\lambda_1I_n\right)^{-1}D,
\end{split}
\end{equation} 
where $I_n$ represents the $n \times n$ identity matrix. For a testing probe sample $s^t$ and its local matching score vector $\mathbf{z}^t$, the global matching score vector $y^t$ is obtained by $y^t=\mathbf{z}^tW$. For a kernel version, it is obtained by:
\begin{equation}\label{wk1}
\begin{split}
y^t_{k} & =\Phi(\mathbf{z}^t)W_{k}  \\ 
&   =\Phi(\mathbf{z}^t)\Phi^T \left(\Phi\Phi^T+\lambda_1I_n\right)^{-1}D  \\
& =K(\mathbf{z}^t,\mathbf{z})\left(K(\mathbf{z},\mathbf{z})+\lambda_1I_n\right)^{-1}D,\\
\end{split}
\end{equation}
where \mbox{$K(\mathbf{z}^t,\mathbf{z})=[k(\mathbf{z}^t,\mathbf{z}^1),k(\mathbf{z}^t,\mathbf{z}^2),...,k(\mathbf{z}^t,\mathbf{z}^n)]$} and \mbox{$K(\mathbf{z},\mathbf{z})=\{k(\mathbf{z}_i,\mathbf{z}_j)\}$} are both kernel computations. 

One potential disadvantage of the above kernel model is its scalability. During the training phase, the complexity of computing and storing $K(\mathbf{z},\mathbf{z})$ is significant for large size problems. Therefore, a random sample-selection technique introduced in Zhang \textit{et al.} \cite{Zhang201789} can be applied to reduce the kernel complexity of large-scale datasets. The assumption behind this is to select a small number of samples that could represent the distribution of a large-scale dataset. If $n_k (n_k \ll n)$ samples are selected from the probe set for the kernel model, this reduces the kernel complexity from $O(n \times n)$ to $O(n_k \times n_k)$.

\subsection{Global matching combination}

After the global matching score matrix of the probe set is obtained. it can be used to update the local matching identity matrix and obtain the global matching identity matrix $G=\{g_{ij}\}$, with size $n\times m$, where each element $g_{ij}$ represents the global matching identity of the $i^{th}$ sample's $j^{th}$ patch. The rule is to apply a threshold $t$ on the value of global matching score. The motivation is that if the global matching of one patch is small than the threshold, the corresponding global matching identity will be ignored in the global matching combination:
\begin{equation}\label{out2}
g_{ij} = \left\{ \begin{array}{rcl}
p_{ij}   & y_{ij} >= t \\
-1  &  y_{ij} < t \\
\end{array}\right..
\end{equation}

\begin{figure} 
	\centering
	\begin{center}
		\includegraphics[width=1\linewidth]{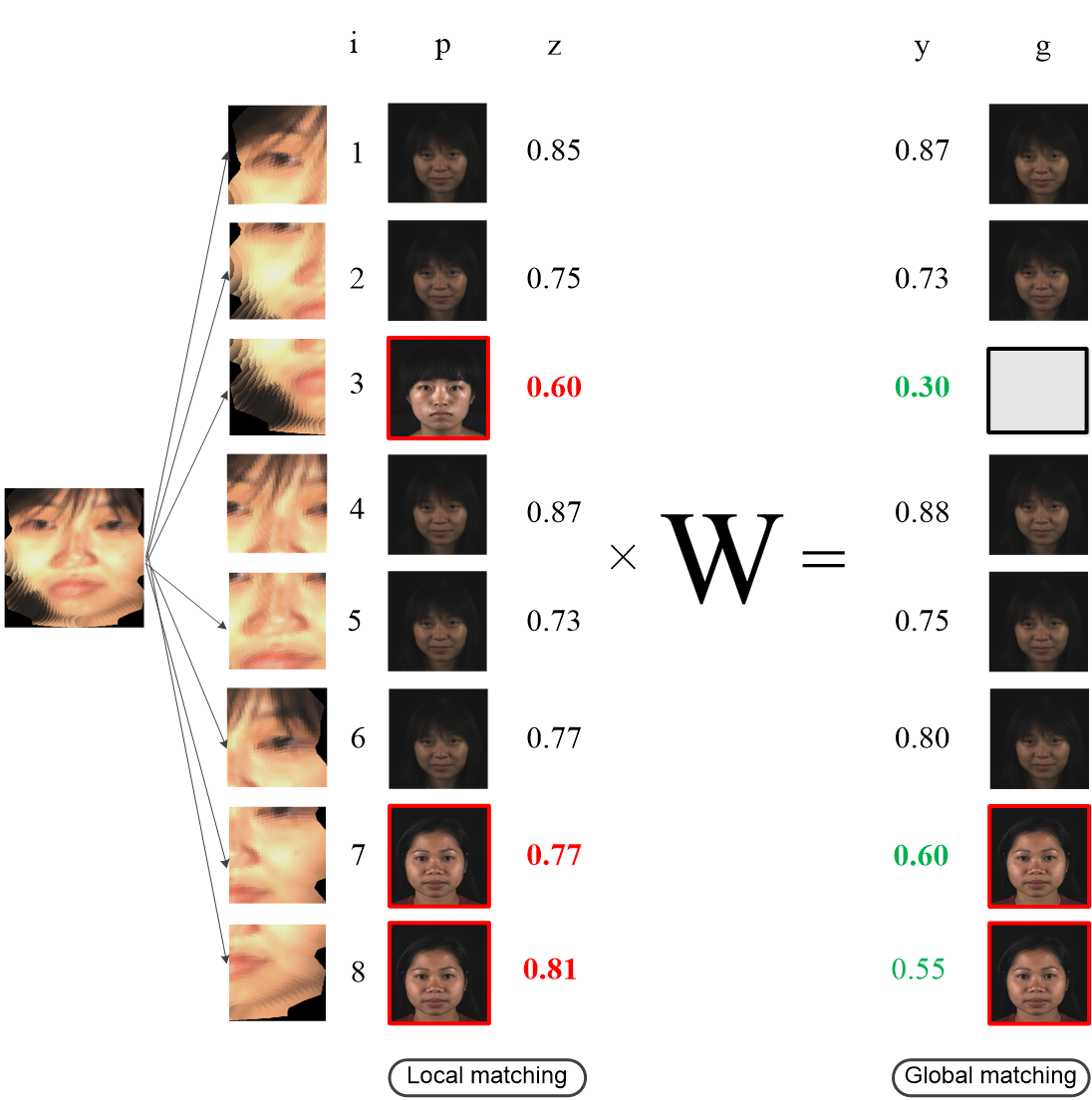}
	\end{center}
	\caption{An example shows the relationship between local matching and global matching. The patch number is represented by $i$. The local matching identity vector and score vector are represented by $p$ and $z$, respectively. The global matching identity vector and score vector are represented by $y$ and $g$, respectively. It can be observed from the local matching identity, several patches are misclassified indicated with red box in $p$ and bold in $z$. After applying the fully associative matrix, the global matching scores are improved significantly. The scores of the misclassified patches are decreased indicated with green and bold in $y$ ($y_3$, $y_7$, and $y_8$). Also, the incorrect global matching identity $g_3$ is also ignored (indicated with black box in $g$) based on that its score value is lower than a learned value $t=0.4$.}
	\label{pipeline2}
\end{figure}

Figure~\ref{pipeline2} depicts an example of the relationship between local matching and global matching. The example shows that the learned matrix can refine the global matching results of all the patches and locate the incorrect global matching identity. 

After the global matching identity matrix and score matrix are obtained. The $\ell_1$-regularized weighting is applied to combine the global matching identities of all the patches. The intuition is to learn different weights for different global matching identities based on their patch locations. Let $\mathbf{q}=\{q_1, q_2,\cdots, q_m\}^T$ represent the weight vector for different patches, and $\sum{^m_{i=1}{q_i}=1}$. Following Zhu \textit{et al.} \cite{zhu2012multi}, a decision matrix \mbox{$H =\{h_{n,m}\}\in {{\mathbb R}}^{n\times m}$ } is defined as:
\begin{equation} \label{GrindEQ__2_} 
h_{i,j}=\left\{ \begin{array}{c}
+1,\ \ \ \emph{if}  \ \ g_{ij} = c_i\  \\
-1,\ \ \ \emph{if}  \ \ g_{ij} \neq c_i
\end{array}.
\right.  
\end{equation}
Note that $h_{i,j}=1$ means that  $g_{ij}$ gives a correct matching, otherwise it gives a incorrect matching. To measure the mismatching of all the patches, the ensemble margin of the $i^{th}$ sample can be defined as:
\begin{equation} \label{GrindEQ__3_} 
\varepsilon \left(s_i \right)=\sum^{m}_{j=1}{q_jh_{ij}}.                                                                    
\end{equation} 
For the probe set $\mathcal{S}$, the ensemble loss under square loss can be defined as:
\begin{equation} \label{GrindEQ__4_} 
\begin{split} 
Loss\left({\mathcal{S}}\right) & =\sum^n_{i=1}{{\left[1-\varepsilon \left(s_i\right)\right]}^2}   \\
& =\sum^n_{i=1}{{\left(1-\sum^{m}_{j=1}{q_jh_{ij}}\right)}^2}\\
& ={\left\|\mathbf{e}-H\mathbf{q}\right\|}^2_2,                        
\end{split} 
\end{equation} 
where $\mathbf{e}=\left[ 1, 1, \cdots, 1 \right]^T$, and $dim(\mathbf{e})=m$. Considering that some patches do not make much contribution, sparsity of $\mathbf{q}$ with the $\ell_1$-norm is introduced. Also the learned weights should be positive. With these constraints, the optimization problem becomes:
\begin{equation} \label{GrindEQ__5_} 
\begin{array}{c}
{\left\|\mathbf{e}-H\mathbf{q}\right\|}^2_2+{\lambda_2 \left\|\mathbf{q}\right\|}_1 \\ 
s.t.\ \sum^{m}_{i=1}{q_i}=1, q_i>0,\ i=1, 2, \cdots, m. \end{array}                                   
\end{equation} 
Using the same strategy as Zhu \textit{et al.} \cite{zhu2012multi}, converting the weight constraint to $\mathbf{e}\mathbf{q}=1$, and adding to the objective function, the function becomes: 
\begin{equation} \label{GrindEQ__6_} 
\begin{array}{c}
\mathbf{q}^* ={argmin}_{\mathbf{q}}\{{\left\|\mathbf{e}'-H'\mathbf{q}\right\|}^2_2+{\lambda_2 \left\|\mathbf{q}\right\|}_1\} \\ 
s.t.\ q_i>0,\ i=1,\ 2,\ \cdots, m \end{array},                                  
\end{equation} 
where $\mathbf{e}'=\left[\mathbf{e};1\right]$, $H'=\left[{H};\mathbf{e}^T\right]$.  The function can be solved using popular $\ell_1$-minimization methods. After weight learning, for a testing probe sample $s^t$, the final matching identity is $u^t$ obtained by 
\begin{equation} \label{GrindEQ__7_} 
u^t=\arg max_c \{ \sum{q_iy_i|g^t_{i}=c \} }.
\end{equation} 
 Thus, the global matching score and weight information are combined to obtained the final matching identity and score. The matching score of $s$ is also taken into account in the final matching with weight of 1. The proposed method for signature matching is summarized in Algorithm \ref{a2}. The relationships and differences between different patches are well explored. This information is ignored in the current UR2D system.      

  \begin{algorithm}
	\caption{Signature matching}
	\label{a2}
	\KwIn{Gallery signature list $\{\mathbb{S}^g_i\}$, probe image signature $\mathbb{S}^p=\{E^p, O^p\}$ and $t$}
	\KwOut{final matching identity $u$}
	Compute local matching identity vector $p$ and score vector $z$ \\
	Compute global matching score vector $y$ based on Eq.(\ref{wk1})  \\
	Compute global matching identity vector $g$ based on Eq.(\ref{out2}) \\
	Compute final matching $u$ label based on Eq.(\ref{GrindEQ__7_})\\
	\Return{\{u\}};
\end{algorithm}

\section{Experiments}
\label{sec5}
This section presents the evaluation of the proposed FAPSM matcher on two types of face recognition scenarios: a constrained environment and an unconstrained environment. The datasets used for testing are the UHDB31 dataset \cite{ha2017uhdb31} and the IJB-A dataset \cite{klare2015pushing}, respectively. The latest UR2D is used as a baseline pipeline based on PRFS and DPRFS signatures. A training set with 1,000 identities based on the CASIA WebFace dataset \cite{yi2014learning} is created. The overlapping identities with the IJB-A dataset are removed before the selection. Then, the training set is divided equally into two sets as gallery and probe set. Following Xu \textit{et al.} \cite{xiang2017ijcb}, the results of VGG-Face, FaceNet, and COTS v1.9 are also used for comparison. The threshold $t$ is set to 0.4 in the range of $\{0.2, 0.3, ..., 0.6\}$. Gaussian kernel ($\sigma=0.05$) is used in the proposed matcher with $\lambda_1 = 1$. The parameters are learned from the training probe set. The Rank-1 accuracy is used as performance measurement.

\subsection{Constrained face recognition}
\begin{figure} 
     \centering
   \begin{center}
     \includegraphics[width=1\linewidth]{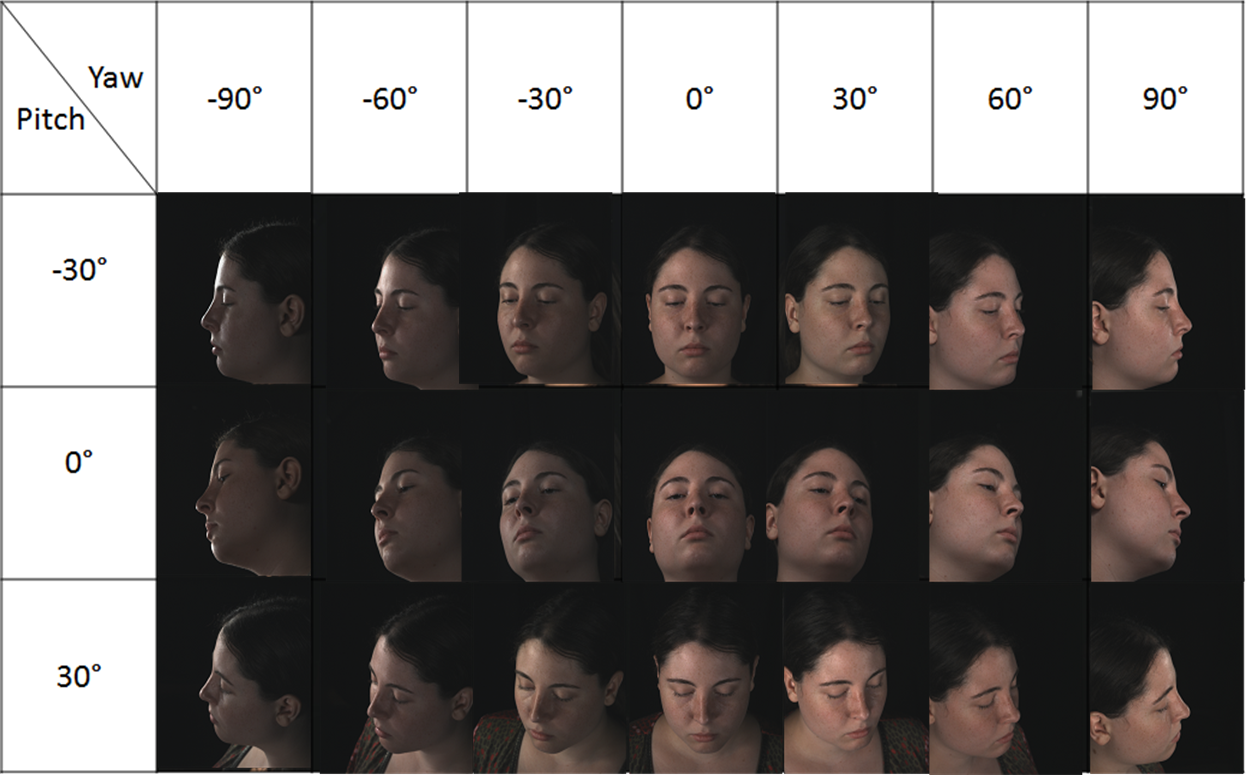}
   \end{center}
      \caption{Image examples of different poses in the UHDB31 dataset.}
   \label{uhdb31_ex}
   \end{figure}

The UHDB31 dataset \cite{ha2017uhdb31} contains 29,106 color face images of 77 subjects with 21 poses and 18 illuminations. To exclude the illumination changes, a subset with nature illumination is selected. To evaluate the performance of cross pose face recognition, the frontal-pose face images are used as gallery and the remaining images from 20 poses are used as separate probe sets. Figure~\ref{uhdb31_ex} shows the example images from different poses. Table \ref{uhdb31_r1} depicts the performance of different methods. It can be observed that the proposed FAPSM matcher can improve the accuracy under five poses, especially some large poses like pose-1 to pose-3 and pose-19 to pose-21. The accuracy improvements range from 1\% to 3\%. At the same time, the excellent performance of the close-to-frontal poses is retained.       

\begin{table*}
	\begin{center}
	\caption{The Rank-1 performance of different methods on the UHDB31 dataset (\%). The methods are ordered as VGG-Face, COTS v1.9, FaceNet, UR2D-PRFS, UR2D-DPRFS, and FAPSM.}
	\vspace{-7px} 
		\scalebox{1}{ 
	\begin{tabular}{| c|c |c| c |c |c| c| c|} 
	\hline 
	
    \backslashbox{Pitch}{Yaw}
    & -90\textdegree{} &-60\textdegree{} &-30\textdegree{} &0\textdegree{} & +30\textdegree{} &+60\textdegree{} &+90\textdegree{} \\
    \hline
    +30\textdegree{} & 
    \makecell{14,11,58, \\ 48,82,{\bf 83}} & 
    \makecell{69,32,95, \\ 90,{\bf 99},{\bf 99}} & 
    \makecell{94,90,{\bf 100},  \\ {\bf 100},{\bf 100},{\bf 100}} & 
    \makecell{99,{\bf 100},{\bf 100}, \\ {\bf 100},{\bf 100},{\bf 100}} & 
    \makecell{95,93,99, \\ {\bf 100},99,99} & 
    \makecell{79,38,92, \\ 95,{\bf 99},{\bf 99}} & 
    \makecell{19,7,60, \\ 47,75,{\bf 78} }   \\
        \hline
    0\textdegree{} & 
    \makecell{22,9,84, \\ 79,96,{\bf 97}} & 
    \makecell{88,52,99, \\ {\bf 100},{\bf 100},{\bf 100}} & 
    \makecell{{\bf 100},99,{\bf 100},  \\ {\bf 100},{\bf 100},{\bf 100}} & - & 
    \makecell{{\bf 100},{\bf 100},{\bf 100}, \\ {\bf 100},{\bf 100},{\bf 100}} & 
    \makecell{94,73,99, \\ {\bf 100},{\bf 100},{\bf 100}} & 
    \makecell{27,10,91, \\ 84,{\bf 96},{\bf 96}}    \\
        \hline
        -30\textdegree{} & 
    \makecell{8,0,44, \\ 43,75,{\bf 76}} & 
    \makecell{2,19,80, \\ 90,{\bf 97},{\bf 97}} & 
    \makecell{91,90,99,  \\ 99,{\bf 100},{\bf 100}} & 
    \makecell{96,99,99, \\ {\bf 100},{\bf 100},{\bf 100}} & 
    \makecell{96,98,97, \\ 99,{\bf 100},{\bf 100}} & 
    \makecell{52,15,90, \\ 95,{\bf 96},95} & 
    \makecell{9,3,35, \\ 58,{\bf 79},{\bf 79}}    \\
    \hline
	\end{tabular}}
	\label{uhdb31_r1}
	\end{center}
\end{table*}

\subsection{Unconstrained face recognition}
The IJB-A dataset \cite{klare2015pushing} contains images and videos from 500 subjects captured from the ``in the wild'' environment. This dataset merges images and frames and provides evaluations on the template level. A template contains one or several images/frames of one subject. According to the IJB-A protocol, it splits galleries and probes into 10 splits. In this experiment, the same modification as Xu \textit{et al.} \cite{xiang2017ijcb} is followed for use in close-set face recognition. The performance of different methods is shown in Table \ref{ijba_r1}. The performance of FaceNet is ignored as its training set contains overlapping identities with the IJB-A dataset.

\begin{table*}
	\begin{center}
	\caption{The Rank-1 performance of different methods on the IJB-A dataset (\%).}
	\vspace{-7px} 
		\scalebox{0.9}{ 
	\begin{tabular}{ l |c c c c c c c c c c c} 
	\hline 
	Methods &split-1 & split-2 &split-3 &split-4 &split-5 & split-6 & split-7 &split-8 & split-9 & split-10 & Average\\
		\hline 
    VGG-Face &76.18 &74.37 &24.33 &47.67 &52.07 &47.11 &58.31 &54.31 &47.98 &49.06 &53.16 \\
    COTS v1.9 &75.68 &76.57 &73.66 &76.73 &76.31 &77.21 &76.27 &74.50 &72.52 &77.88 &75.73\\
    	\hline 
    	    UR2D-PRFS &49.01	&49.57	&48.22	&47.75	&48.85	&44.46	&52.46	&48.22	&43.48	&48.79	&48.08\\
    UR2D-DPRFS &78.78&77.60&	77.94&	79.88&	78.44&	80.57&	81.78&	79.00&	75.94&	79.22&	78.92 \\
      FAPSM &{\bf 79.38} &{\bf 78.17}& 	{\bf 78.83} & {\bf 80.42} &{\bf 79.33}&{\bf 81.00}&{\bf 82.29}&{\bf 79.20}&	{\bf 76.55}&{\bf 79.30}&{\bf 79.47} \\

	\hline 
	\end{tabular}}
	\label{ijba_r1}
	\end{center}
\end{table*}

%
%

From Table \ref{ijba_r1}, it can be observed that the FAPSM matcher achieves better performance under all the splits. The accuracy is improved on average by 0.55\%. Overall, the proposed matcher achieves the best result on all the splits compared to VGG-Face, COTS v1.9, and UR2D. The reason behind this is that with fully associative learning, the proposed matcher can improve the matching of different patches. With global matching combination, the final matching is more robust than previous methods. The proposed matcher also works for UR2D-PRFS with improvements, however, the performance is still worse than that of the CNN based DPRFS signature. 

\subsection{Discussion}

\begin{figure*}
	\centering
	\begin{subfigure}[b]{0.45\textwidth}
		\includegraphics[width=\textwidth]{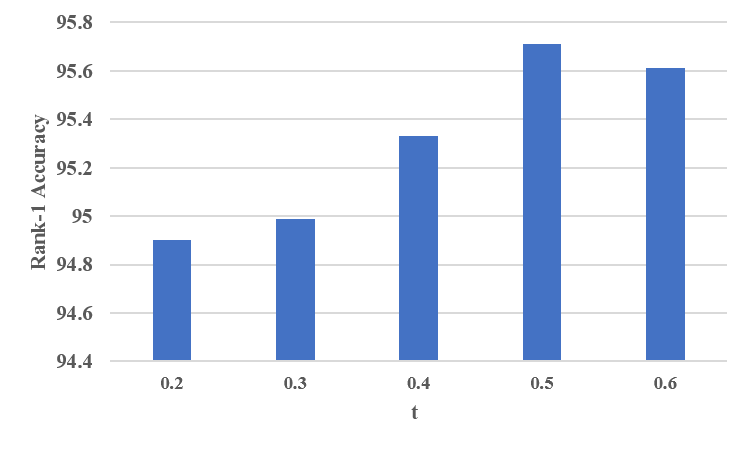}
		\caption{ }
		\label{fig:gull}
	\end{subfigure}%
	\begin{subfigure}[b]{0.45\textwidth}
		\includegraphics[width=\textwidth]{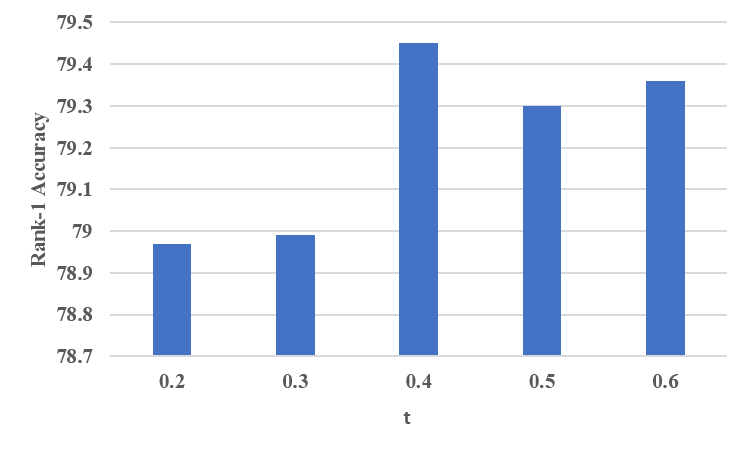}
		\caption{ }
		\label{fig:tiger}
	\end{subfigure}
	\caption{ The sensitivity of $t$ in FAPSM. (a) UHDB31. (b) IJB-A.}\label{re-4-p-f1}
\end{figure*}

The sensitivity of the value of $t$ is analyzed in the set of \{0.2, 0.3, 0.4, 0.5, 0.6\}. The results are shown in Figure~\ref{re-4-p-f1}. From Figure~\ref{re-4-p-f1}, it can be observed that different performance is obtained with threshold values. the best value learned from the training set is used in previous experiments.

The statistical analysis for FAPSM and UR2D-DPRFS (the best baseline) is also performed over the 30 data splits (20 from UHDB31 and 10 from IJB-A). Following Dem\v{s}ar \textit{et al.} \cite{demvsar2006statistical},  \cite{dunn1961multiple} are used to compare multiple methods over multiple datasets. Let $r_i^j$ represent the rank of the $j^{th}$ of k algorithm on the $i^{th}$ of $N$ datasets.  The Friedman test compares the average ranks of different methods, by $R_j = \frac{1}{N} \sum_i r_i^j$. The null-hypothesis states that all the methods are equal, so their ranks $R_j$ should be equivalent. The original Friedman statistic \cite{friedman1937use,friedman1940comparison}, 
\begin{equation}\label{st1}
\mathcal{X}_F^2 = \frac{12N}{k(k+1)}[\sum_j R_j^2 - \frac{k(k+1)^2}{4}],
\end{equation}
is distributed according to $\mathcal{X}_F^2$ with $k-1$ degrees of freedom. Due to its undesirable conservative property,  Iman \textit{et al.} \cite{iman1980approximations} derived a better statistic
\begin{equation}\label{st2}
F_F = \frac{(N-1)\mathcal{X}_F^2}{N(k-1)-\mathcal{X}_F^2},
\end{equation}
which is distributed according to the F-distribution with $k-1$ and $(k-1) \times (N-1)$ degrees of freedom. First, the average rank of each method is computed as 1.28 and 1.72 for FAPSM and UR2D-DPRFS, respectively. The $F_F$ statistical value of the Rank-1 accuracy is computed as $6.01$. With two methods and 30 data splits, $F_F$ is distributed with $2-1$ and $(2-1) \times (30-1) = 29$ degrees of freedom. The critical value of $F(1, 29)$ for $\alpha = 0.10$ is $2.89 < 6.01$, so the null-hypothesis is rejected. Then, the two tailed Bonferroni-Dunn test is applied to compare the two methods by the critical difference:
\begin{equation}\label{st3}
CD = q_{\alpha} \sqrt{\frac{k(k+1)}{6N}},
\end{equation}
where $q_{\alpha}$ is the critical values. If the average rank between two methods is larger than the critical difference, the two methods are significantly different. The critical value of two methods when $p = 0.10$ is 1.65. the critical difference is computed as $CD = 1.65 \sqrt{\frac{2 \times 3}{6 \times 30}} = 0.30$. Then in conclusion, under the Rank-1 accuracy, FAPSM performs significantly better than UR2D-DPRFS (the difference between their ranks is $1.72 - 1.28 = 0.44 > 0.30$).

\section{Conclusion}
\label{sec6}
This paper proposed a patch-based 1-to-N signature matcher method for face recognition that learns the correlations between different facial patches. A weight matrix was learned to update the local matching identity of each patch and obtain the global identity. The global identities of all the patches were combined to obtain the final matching identity. The experimental results confirmed the assumption that the learned correlations can be used to improve matching performance. Compared to the UR2D system, the Rank-1 accuracy was improved by 3\% for the UHDB31 dataset and 0.55\% for the IJB-A dataset. The limitation of the current matcher is that it is trained with fixed patch division. The influence of patch division on the fully associative learning will be investigated in the future.

 \section*{Acknowledgements}
This material is based upon work supported by the U.S. Department of Homeland Security under Grant Award Number 2015-ST-061-BSH001. This grant is awarded to the Borders, Trade, and Immigration (BTI) Institute: A DHS Center of Excellence led by the University of Houston, and includes support for the project ``Image and Video Person Identification in an Operational Environment: Phase I'' awarded to the University of Houston. The views and conclusions contained in this document are those of the authors and should not be interpreted as necessarily representing the official policies, either expressed or implied, of the U.S. Department of Homeland Security.

{\small
\bibliographystyle{ieee}
\bibliography{egbib_all_p}
}

\end{document}